\documentclass[sigconf]{acmart}
\renewcommand\footnotetextcopyrightpermission[1]{}
\AtBeginDocument{%
  }

\setcopyright{acmlicensed}
\copyrightyear{2018}
\acmYear{2018}
\acmDOI{XXXXXXX.XXXXXXX}
\acmConference[MM ’25]{ACM Multimedia 2025}{October 27--31, 2025}{Dublin, Ireland}
\acmISBN{978-1-4503-XXXX-X/2018/06}

\acmSubmissionID{564}

\usepackage{amssymb} 
\usepackage{bbding}
\usepackage{graphicx}
\usepackage{booktabs}
\usepackage{array}
\usepackage{multirow}
\usepackage{makecell}
\usepackage{caption}
\usepackage{paracol}

\begin{document}
\begin{sloppypar}

\title{HairShifter: Consistent and High-Fidelity Video Hair Transfer via Anchor-Guided Animation}

\author{Wangzheng Shi}
\authornote{Equal contribution.}
\affiliation{%
  \institution{Xiamen University}
  \department{School of Informatics}
  \city{Xiamen}
  \country{China}
}
\email{shiwangzheng@stu.xmu.edu.cn}

\author{Yinglin Zheng}
\authornotemark[1] 
\affiliation{%
  \institution{Xiamen University}
  \department{School of Informatics}
  \city{Xiamen}
  \country{China}
}
\email{zhengyinglin@stu.xmu.edu.cn}

\author{Yuxin Lin}
\affiliation{%
  \institution{Xiamen University}
  \department{School of Informatics}
  \city{Xiamen}
  \country{China}
}
\email{linyx@stu.xmu.edu.cn}

\author{Jianmin Bao}
\affiliation{%
  \institution{Microsoft Research Asia}
  \city{Beijing}
  \country{China}
}
\email{jianbao@microsoft.com}

\author{Ming Zeng}
\authornote{Corresponding authors.}
\affiliation{%
  \institution{Xiamen University}
  \department{School of Informatics}
  \city{Xiamen}
  \country{China}
}
\email{zengming@xmu.edu.cn}

\author{Dong Chen}
\affiliation{%
  \institution{Microsoft Research Asia}
  \city{Beijing}
  \country{China}
}
\email{doch@microsoft.com}

\renewcommand{\shortauthors}{Wangzheng Shi, Yinglin Zheng et al.}

\begin{abstract}
 Hair transfer is increasingly valuable across domains such as social media, gaming, advertising, and entertainment. While significant progress has been made in single-image hair transfer, video-based hair transfer remains challenging due to the need for temporal consistency, spatial fidelity, and dynamic adaptability. In this work, we propose HairShifter, a novel "Anchor Frame + Animation" framework that unifies high-quality image hair transfer with smooth and coherent video animation. At its core, HairShifter integrates a Image Hair Transfer (IHT) module for precise per-frame transformation and a Multi-Scale Gated SPADE Decoder to ensure seamless spatial blending and temporal coherence. Our method maintains hairstyle fidelity across frames while preserving non-hair regions. Extensive experiments demonstrate that HairShifter achieves state-of-the-art performance in video hairstyle transfer, combining superior visual quality, temporal consistency, and scalability. The code will be publicly available. We believe this work will open new avenues for video-based hairstyle transfer and establish a robust baseline in this field.
\end{abstract}


\begin{teaserfigure}
  \includegraphics[width=\textwidth]{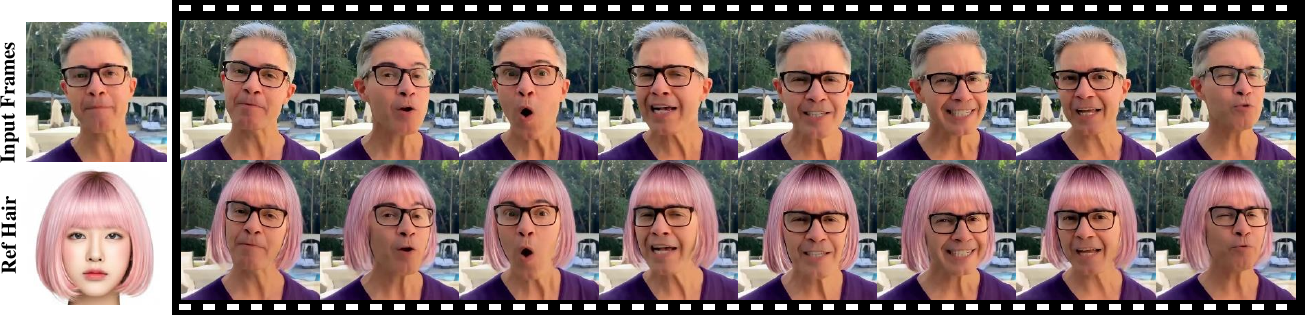}
  \caption{High-Fidelity Video Hairstyle Transfer with HairShifter. We successfully transfers the reference hairstyle while precisely preserving the subject's identity, expression, and background details across dynamic motion.}
  \Description{Enjoying the baseball game from the third-base
  seats. Ichiro Suzuki preparing to bat.}
  \label{fig:teaser}
\end{teaserfigure}

\received{20 February 2007}
\received[revised]{12 March 2009}
\received[accepted]{5 June 2009}

\maketitle

\vspace{-0.75em}
\section{Introduction}
Hair Transfer~(also called Hair Replacement) is an important task in the area of Intelligent Multimedia Processing. It has numerous applications in industries such as social media, gaming, advertising, film, and television. 

Currently, a serial of works on single-image hair transfer\cite{zhu2021barbershop,kim2022style,wei2022hairclip,wei2023hairclipv2,nikolaev2024hairfastgan,zhang2024stable,zeng2024hairdiffusion} achieve satisfactory results. For example, the methods based on Generative Adversarial Networks (GANs)\cite{tan2020michigan,wei2022hairclip,wei2023hairclipv2,isola2017image,karras2019style,zhu2021barbershop,khwanmuang2023stylegan,kim2022style,nikolaev2024hairfastgan}, and the methods based on Latent Diffusion Models(LDMs)\cite{zhang2024stable,zeng2024hairdiffusion,ho2020denoising,yang2023uni,zhuang2024task,rombach2022high}. Despite these progress, extending these successes to video remains a significant challenge. We analyze potential approaches for video hair transfer and their limitations.

A basic approach for video hair transfer involves applying single-image methods frame by frame. However, this naive frame-wise transfer lacks temporal coherence, resulting in visible flickering, structural jittering, and instability in non-hair areas. Alternatively, while modern text-to-video (T2V) generation methods \cite{wu2023tune, qi2023fatezero, ceylan2023pix2video, geyer2023tokenflow, cong2023flatten, jeong2023ground} can model temporal dependencies, their open-domain nature and limited fine-grained control hinder precise hair transfer and identity preservation. Furthermore, their applicability is often restricted by sequence length limitations.

Video editing models \cite{bai2024uniedit, chai2023stablevideo, yang2025videograin} offer some promise for preserving background regions by using masks for inpainting. However, their effectiveness is limited by the practical challenge of obtaining accurate auxiliary masks. Furthermore, portrait animation techniques \cite{wei2024aniportrait, yang2024megactor, xie2024x, guo2024liveportrait}, while capable of generating temporally smooth sequences through motion transfer, inherently deform the entire image. This global deformation makes them ill-suited for the selective replacement required for hair transfer, as it inevitably leads to unwanted modifications in non-hair areas.

We summarize these methods and their properties in Table \ref{tab:video_transfer_comparison}, revealing none of these existing paradigms fully satisfy the core requirements for ideal video hair transfer: \textit{ Temporal Consistency}, \textit{Non-Hair Region Fidelity}, \textit{Hair Transfer Quality}, and \textit{Frame Scalability}.

To address these challenges, we propose an "Anchor Frame + Animation" pipeline. This approach combines high-fidelity image hair transfer with animation's temporal smoothness in two key stages. First, we create a high-quality "anchor frame" using image hair transfer. This static image embodies the desired hairstyle and the identity of the driving video subject. The anchor frame then serves as the consistent hair source. Second, we animate this anchor frame using motion cues from the driving video. This process ensures temporally coherent hair transfer, dynamically applying the hairstyle while preserving identity and expressions throughout the video.

\begin{table}[t]
  \centering
  \resizebox{1\columnwidth}{!}{ 
  \begin{tabular}{lcccc}
    \toprule
    \textbf{Method} & \textbf{\makecell[c]{Temporal\\Consistency}} & \textbf{\makecell[c]{Non-Hair\\Region Fidelity}} & \textbf{\makecell[c]{Hair Transfer\\Quality}} & \textbf{\makecell[c]{Frame\\Scalability}} \\
    \midrule
    Image Hair Transfer\cite{nikolaev2024hairfastgan,zhang2024stable,zeng2024hairdiffusion} & \XSolidBrush & \Checkmark & \Checkmark & \Checkmark \\
    T2V Generation\cite{wu2023tune, qi2023fatezero, ceylan2023pix2video, geyer2023tokenflow}      & \Checkmark   & \XSolidBrush & \XSolidBrush & \XSolidBrush \\
    Video Editing\cite{bai2024uniedit, chai2023stablevideo, yang2025videograin}       & \Checkmark   & \XSolidBrush & \XSolidBrush & \XSolidBrush \\
    Portrait Animation\cite{wei2024aniportrait, yang2024megactor, xie2024x, guo2024liveportrait}     & \Checkmark   & \XSolidBrush & \XSolidBrush & \Checkmark \\
    \textbf{HairShifter~(Ours)}       & \Checkmark   & \Checkmark & \Checkmark & \Checkmark \\
    \bottomrule
  \end{tabular}
  } 
  \caption{Comparison of Video Hair Transfer Paradigms. Only our method satisfies all core requirements.}
  \label{tab:video_transfer_comparison}
\end{table}
\vspace{-0.75em}

To enhance anchor frame quality, we utilize an Image Hair Transfer (IHT) module. IHT is based on the state-of-the-art image hair transfer model (Stable-Hair \cite{zhang2024stable}) but with better robustness to pose variations, thereby generating high-quality hair-swapped anchor frame.

Even with high-quality anchor frames, the `Anchor + Animation'' strategy faces challenges in preserving non-hair regions and achieving harmonious blending. To tackle this, we propose a decoupled training strategy and a Multi-Scale Gated SPADE Decoder. The decoupled training strategy uses IHT to generate pseudo driving frames, compelling the network to learn hair appearance from the source frame and non-hair features from the pseudo driving frame. The MSG-SPADE Decoder employs a dual-pathway architecture and multi-scale gated fusion to achieve seamless blending and enhance spatio-temporal consistency. 

Our framework, named \textit{HairShifter}, balances high-quality hair generation, dynamic adaptability, temporal continuity, non-hair region preservation, and seamless blending. It is efficient, has no length limitations, and achieves state-of-the-art performance, as demonstrated in Figure \ref{fig:teaser} and our experiments.

In summary, our contributions are outlined as follows:
\begin{itemize}
\item To the best of our knowledge, we propose the first framework specifically designed for video hairstyle transfer, effectively balancing generation quality, temporal consistency, and preservation/blending fidelity, achieving state-of-the-art results.
\item We design a novel training strategy that repurposes an image hair transfer module to decouple appearance and motion, resolving the core conflict between maintaining temporal consistency and preserving non-hair region details.
\item We develop the MSG-SPADE Decoder, whose dual-path, reference-guided architecture with gated fusion enhances identity retention and achieves seamless spatial integration for temporally stable video synthesis.
\end{itemize}

\section{Related Work}

\noindent \textbf{Hair Transfer.}
Early GAN-based hair transfer focused on editable synthesis via latent space manipulation, such as attribute decomposition\cite{tan2020michigan}, preservation\cite{zhu2021barbershop}, and style-space modification\cite{wu2022hairmapper,kim2022style,wei2022hairclip,khwanmuang2023stylegan}. Subsequent works enhanced control and efficiency, with HairCLIPV2\cite{wei2023hairclipv2} enabling text-driven editing and HairFastGAN\cite{nikolaev2024hairfastgan} achieving fast, pose-robust generation. Recently, diffusion models\cite{ho2020denoising,rombach2022high,dhariwal2021diffusion}, often guided by spatial constraints like ControlNet\cite{zhang2023adding}, have gained prominence. Methods like Stable-Hair and HairDiffusion\cite{zeng2024hairdiffusion} improved attribute alignment and fine-grained control. However, these techniques are fundamentally designed for static images and lack the temporal modeling.

\noindent \textbf{Portrait Animation.}
Existing portrait animation methods animate source images using motion priors, such as keypoints or flow fields\cite{siarohin2019first,mallya2022implicit,zhao2022thin,hong2022depth,hong2023implicit,zeng2023face,han2024face,wei2024aniportrait,drobyshev2024emoportraits}. While advanced techniques have achieved photorealistic rendering\cite{wang2021one}, enhanced spatiotemporal coherence\cite{yang2024megactor}, performed cross-identity synthesis\cite{xie2024x}, and captured subtle facial dynamics\cite{guo2024liveportrait}, they typically operate holistically on the entire image. This lack of part-specific control renders them unsuitable for tasks like hairstyle transfer that demand region-aware identity preservation.

\noindent \textbf{Video Editing.}
Text-to-video (T2V) generation\cite{wu2023tune,qi2023fatezero,ceylan2023pix2video,geyer2023tokenflow,yu2023magvit,guo2023animatediff,xu2024easyanimate} and editing frameworks\cite{yang2023rerender,cong2023flatten,zhang2023controlvideo,jeong2023ground,gu2024videoswap} struggle with fine-grained spatial control, often hampered by a reliance on accurate masks. Even advanced methods that enhance spatiotemporal locality, such as UniEdit\cite{bai2024uniedit}, StableVideo\cite{chai2023stablevideo}, and VideoGrain\cite{yang2025videograin}, do not offer robust mechanisms for targeted manipulation. This architectural deficiency limits their utility for applications like video hairstyle transfer, which mandate precise, spatially-aware and identity-preserving editing.


\vspace{-0.75em}
\section{Methodology}
\vspace{-0.1em}
\subsection{Framework Overview}
\label{subsec:framework_overview} 
We introduce \textit{HairShifter}, a novel framework conceived for high-fidelity, temporally consistent video hair transfer while meticulously preserving non-hair attributes. At its core, HairShifter employs an \textbf{"Anchor Frame + Animation"} pipeline. 

\begin{figure*}[ht] 
  \centering
  \includegraphics[width=\textwidth]{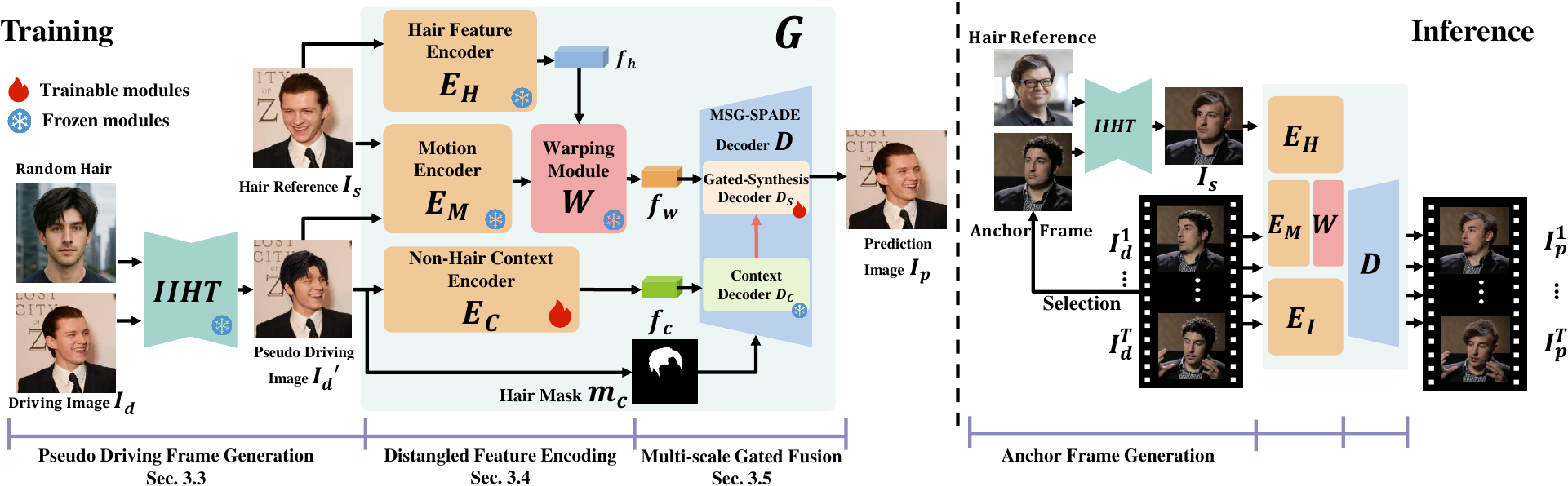}
  \caption{\textbf{Overview of the HairShifter} 
(a) During training, we enforces hair/non-hair decoupling using Pseudo Driving Frame Generation and Disentangled Feature Encoding, with the Multi-Scale Gated Fusion handling seamless feature fusion for synthesis. 
(b) During inference, the Image Hair Transfer(IHT) creates the static anchor frame $I_s$, which is subsequently animated by the animation network $G$ guided by the driving video.}
  \label{fig:pipeline}
\end{figure*}

The inference process, illustrated in Figure \ref{fig:pipeline}(b), involves two main stages. First, a high-quality static anchor image $I_s$ containing the target hairstyle is generated using an \textbf{Image Hair Transfer (IHT)} module. Second, our animation network $G$ synthesizes the output video frame-by-frame by animating this anchor $I_s$, guided by the motion and non-hair context derived from the driving video $V_d = \{I_d^t\}, t\in\{1, ..., T\}$.

To enable this targeted animation during inference, the network $G$ is trained using the specific strategy depicted in Figure \ref{fig:pipeline}(a). This training regime addresses the core challenges: 
(1) \textbf{Forced Decoupling} is achieved via a novel \textbf{training strategy} using pseudo driving frames (Sec \ref{subsec:decoupling_strategy}), forcing the network to separate hair appearance (from source $I_s$) from non-hair dynamics (from driving frame $I_d^t$). 
(2) \textbf{Information Preservation} relies on \textbf{Disentangled Feature Encoding} (Sec \ref{subsec:disentangled_encoding}) to separately capture hair features and essential non-hair context from the driving frame. 
(3) \textbf{Seamless Blending} is handled by our innovative \textbf{Multi-Scale Gated Fusion}(Sec \ref{subsec:multi_scale_gated_spade}), which intelligently fuses these decoupled feature streams at multiple scales for artifact-free synthesis.

The interplay between the high-quality anchor generated by the IHT module (used for anchor generation at inference and pseudo-frame generation during training) and the animation network $G$, trained with these specialized decoupling, preservation, and blending mechanisms, allows HairShifter to effectively navigate the complexities of video hair transfer. Subsequent sections detail these components and strategies.

\subsection{Pseudo Driving Frame Generation}
\label{subsec:decoupling_strategy}

A core challenge in designing the animation network $G$ is enabling controllable video hair transfer during inference. Specifically, the network must learn to synthesize an output frame that combines the desired hair appearance from a static source image $I_s$ with the non-hair dynamics (pose, expression, identity, background) derived from a driving frame $I_d$. Standard image animation training protocols, which typically train the network to reconstruct $I_d$ using $I_s$ and $I_d$ as inputs (often where $I_s$ is derived from $I_d$ or a related frame), are unsuitable for this task. In such setups, the driving frame $I_d$ inherently contains the correct target hair information, providing a "shortcut" that prevents the network from learning to rely exclusively on $I_s$ for hair appearance and $I_d$ solely for non-hair dynamics.

To enforce this crucial decoupling, we introduce a novel training strategy centered around the generation of a \textit{pseudo driving frame}, denoted as ${I_d}^\prime$. This ${I_d}^\prime$ is constructed by taking the original driving frame $I_d$ and replacing its hair with an arbitrary, \textit{incorrect} hairstyle, represented by a random reference image $R_{random}$. Generating this ${I_d}^\prime$ requires a capable image hair transfer mechanism. Crucially, this mechanism must be able to: (1) generate high-fidelity, realistic hairstyles, (2) accurately transfer the style from $R_{random}$ onto the identity and pose present in $I_d$, and (3) be robust to potential variations in pose between $I_d$ and $R_{random}$, a common scenario. 

For this purpose, we utilize a high-fidelity \textbf{Image Hair Transfer(IHT)} module, adapted from existing state-of-the-art diffusion-based methods \cite{zhang2024stable} and fine-tuned on video data to enhance robustness to pose variations (details of IHT can be found in the supplementary material). Using this module, we generate the pseudo driving frame as ${I_d}^\prime = \text{IHT}(I_d, R_{random})$. Thus, ${I_d}^\prime$ retains the essential non-hair dynamics (pose, expression, identity, background) of the original driving frame $I_d$ but presents deliberately incorrect hair information.

The animation network $G$ is then trained to reconstruct the original ground truth frame $I_d$ using the source image $I_s$ (which contains the *correct* target hair) and the pseudo driving frame ${I_d}^\prime$ as inputs: $I_p = G(I_s, {I_d}^\prime) \approx I_d$. This training objective fundamentally changes the learning dynamics. Because the hair information in ${I_d}^\prime$ is explicitly incorrect, the network is forced to disregard it and learn to extract the \textbf{target hair appearance} solely from the source image $I_s$. Simultaneously, since ${I_d}^\prime$ contains the correct \textbf{non-hair information} (pose, expression, identity alignment, background), the network learns to extract these dynamic elements from the driving input. This constraint effectively compels $G$ to master the essential decoupling required for controllable hair transfer: Hair Appearance from $I_s$, Non-Hair Dynamics from the driving input (${I_d}^\prime$ during training, $I_d$ during inference).

It is worth noting that the same robust image hair transfer module used here to generate ${I_d}^\prime$ during training is also employed during the inference stage to create the high-quality static anchor frame $I_s$ that provides the consistent target hair appearance throughout the generated video.

\begin{figure*}[ht] 
  \centering
  \includegraphics[width=\textwidth]{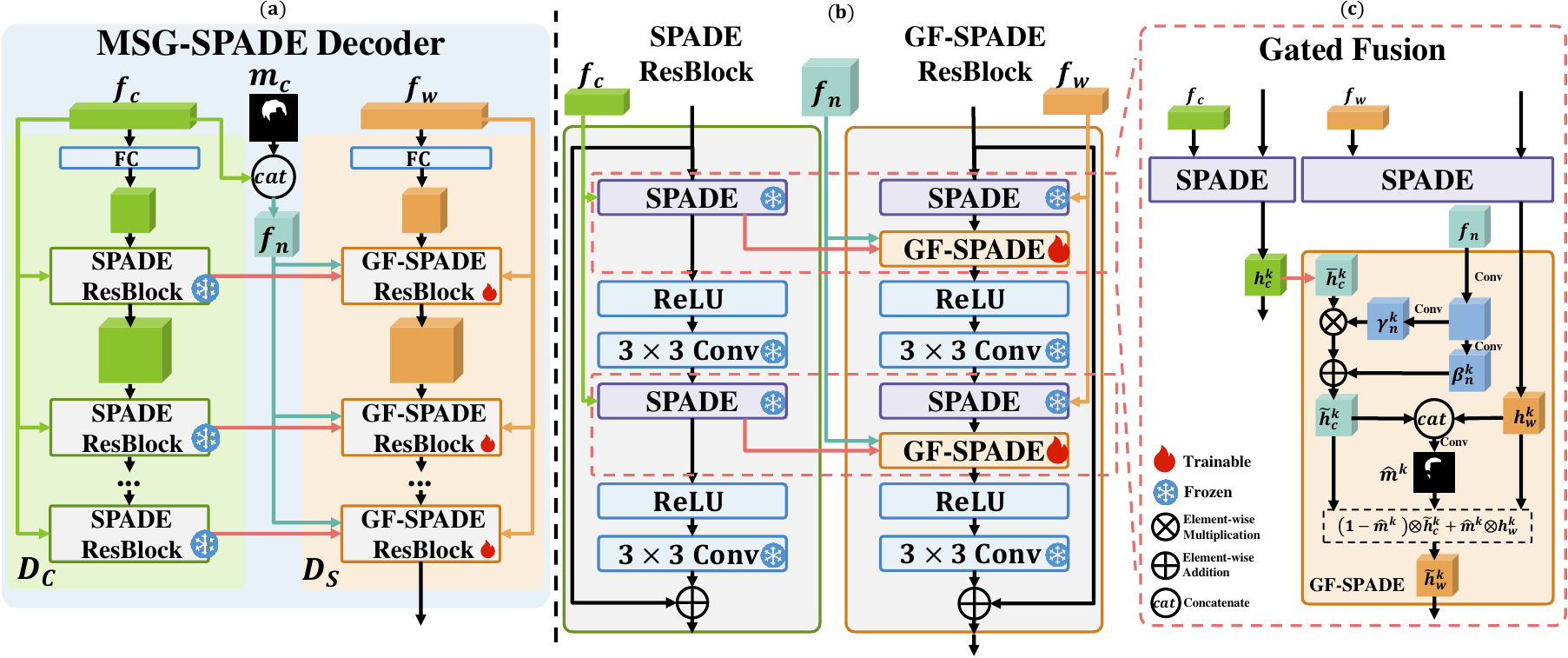} 
\caption{The Multi-Scale Gated SPADE (MSG-SPADE) Decoder. 
(a) Features dual pathways for synthesis ($D_S$, hair $f_w$) and context ($D_C$, non-hair $f_c$). 
(c) The key Gated Fusion (GF-SPADE) blocks within $D_S$ employ hair mask($m_c$) guided modulation and learned gating to fuse features across multiple scales, achieving seamless hair integration while preserving context details.}
  \label{fig:spade}
\end{figure*}

\subsection{Disentangled Feature Encoding}
\label{subsec:disentangled_encoding} 

Our animation network $G$ leverages the robust motion estimation and warping capabilities of the LivePortrait \cite{guo2024liveportrait} framework. Specifically, we utilize its core components: the Motion Estimator ($E_M$), adapt its appearance encoder as our Hair Feature Encoder ($E_H$), and employ its motion-driven feature Warping module ($W$). 

The motion estimator $E_M$ extracts motion representations (capturing pose, expression, etc.) from both the source image $I_s$ and the corresponding driving frame $I_d$ (representing a target time step). Concurrently, the Hair Feature Encoder $E_H$ processes the source image $I_s$ to extract features $f_h$ representing its hair appearance. The warping module $W$ then utilizes the estimated motion difference between the source and driving frame to dynamically transform the hair features $f_h$, producing warped hair features $f_w = W(E_M(I_s), E_M(I_d), E_H(I_s))$. These features $f_w$ effectively carry the target hair appearance information, spatially aligned with the posture and expression depicted in the driving frame.

While $f_w$ provides the hair information, accurately reconstructing the non-hair regions (identity, background, etc.) requires context directly from the driving frame. To ensure meticulous \textbf{Information Preservation} (a core design goal from Sec 3.1), we introduce a dedicated \textbf{Non-hair Context Encoder} $E_C$.

Structurally identical to $E_H$, the encoder $E_C$ processes the driving frame $I_d$ to extract context features $f_c$. These features $f_c$ capture the essential spatial layout, identity cues, and structural details pertaining specifically to the non-hair regions of the target frame $I_d$. This provides the necessary reference for reconstructing everything *except* the hair according to the driving video's content.

Thus, we obtain two primary disentangled feature streams: $f_w$ encoding the motion-aligned hair appearance derived from $I_s$, and $f_c$ encoding the non-hair context derived from $I_d$. Alongside these, the binary hair mask $m_c$ of the driving frame $I_d$ (derived using a face parsing network\cite{zheng2022general}), is also utilized as guidance(\textit{hair mask guidance}). This mask provides explicit spatial guidance regarding the hair region location in the driving frame. All three inputs – the warped hair features $f_w$, the non-hair context features $f_c$, and the hair mask $m_c$ – are then fed into our proposed \textbf{Multi-Scale Gated SPADE Decoder} $D$(detailed in Section \ref{subsec:multi_scale_gated_spade}). This decoder, replacing the standard generator, leverages all inputs to synthesize the final prediction: $I_p = D(f_w, f_c, m_c)$.

\vspace{-0.5em}
\subsection{Multi-Scale Gated SPADE Decoder}
\label{subsec:multi_scale_gated_spade}

A critical challenge in video hair transfer is seamlessly integrating the synthesized target hairstyle while meticulously preserving the identity, expression, and background details present in the driving video's non-hair regions. To address this, we propose the \textbf{Multi-Scale Gated SPADE(MSG-SPADE) Decoder}, a novel architecture designed to replace standard SPADE-based decoders\cite{park2019semantic} commonly found in animation frameworks like LivePortrait\cite{guo2024liveportrait}.

As illustrated in Figure~\ref{fig:spade}(a), the MSG-SPADE Decoder employs a dual-pathway structure to process distinct feature streams. The \textbf{Gated Synthesis Decoder} ($D_S$) primarily operates on the warped feature volume $f_w$, which encodes the target hair appearance dynamically aligned with the driving video's motion. Concurrently, the \textbf{Context Decoder} ($D_C$) processes the context feature volume $f_c$, derived from the driving frame (${I_d}^\prime$ during training, $I_d$ during inference), capturing essential non-hair information such as identity cues, pose, and structural details.

Both pathways are fundamentally constructed using cascaded Residual Blocks, detailed in Figure~\ref{fig:spade}(b), incorporating spatially-adaptive normalization via standard SPADE blocks. The context pathway $D_C$ functions as a standard SPADE decoder, directly utilizing these blocks conditioned on $f_c$. The synthesis pathway $D_S$ also employs standard SPADE blocks, conditioned on $f_w$, but introduces our core innovation: following each standard SPADE block within $D_S$, a Gated Fusion SPADE (GF-SPADE) block is inserted. This GF-SPADE block is specifically designed to fuse the output features from the corresponding standard SPADE block in the context pathway ($D_C$) with the features generated by the preceding standard SPADE block in the synthesis pathway ($D_S$), thereby selectively integrating contextual information at multiple scales.

The core innovation lies in the multi-scale gated fusion mechanism integrated within the GF-SPADE blocks of the synthesis pathway $D_S$. As shown in  Figure~\ref{fig:spade}(c), the output activation for the $k$-th SPADE block in the context pathway and the synthesis pathway are denoted as $h_c^{k}$ and $h_s^{k}$ respectively. Before fusion, the context activation $h_c^{k}$ is further modulated with the guidance of the context features $f_c$ and the hair mask $m_c$ of the driving frame (both concatenated into $f_n$), yielding $h_n^{k}$:
\begin{equation}
\Tilde{h}_c^{k} = \gamma_n^k(f_n) \otimes \bar{h}_c^{k} + \beta_n^k(f_n),
\quad 
f_n=\text{Concat}(f_c, m_c)
\label{eq:fn}
\end{equation}
where $\gamma_n^k$ and $\beta_n^k$ are modulation parameters derived from $f_n$, and $\otimes$ denotes element-wise multiplication. This modulation step helps align the context information precisely before it's blended with the synthesis activation, and incorporating the \textbf{hair mask guidance($m_c$)} provides explicit spatial localization, help improve blending accuracy, particularly at the hair boundaries.

Next, a spatial gating mask $\hat{m}^k$ is computed at each block $k$. This gate adaptively determines the contribution of each pathway at every spatial location. It is derived from the concatenation of the modulated context activation $\Tilde{h}_c^k$ and the synthesis activation $h_w^k$:
\begin{equation}
    \hat{m}^k = \sigma(\text{Conv}(\text{Concat}(h_w^k, \Tilde{h}_c^k)))
\end{equation}
where $\text{Conv}$ represents a convolutional layer and $\sigma$ is the sigmoid activation function, constraining the gate values to the range $[0, 1]$.

Finally, the modulated context activation $\Tilde{h}_c^k$ and the synthesis activation $h_w^k$ are fused using the learned gate $\hat{m}^k$ to produce the output activation $\Tilde{h}_w^{k}$ for the $k$-th GF-SPADE block:
\begin{equation}
    \label{eq:gated_fusion}
    \Tilde{h}_w^{k} = (1 - \hat{m}^k) \otimes \Tilde{h}_c^k + \hat{m}^k \otimes h_w^k
\end{equation}
This output $\Tilde{h}_w^{k}$ then proceeds through the remainder of the GF-SPADE ResBlock (e.g., another convolution, ReLU).

By performing this gated fusion at multiple scales throughout the decoder, the MSG-SPADE architecture effectively integrates the target hairstyle features (via $f_w$ and $h_{w}^k$) with the identity-preserving structural details from the driving frame's context(via $f_c$ and $h_c^k$). The learned gating mechanism ensures that non-hair regions are faithfully reconstructed based on the context pathway, while the hair region smoothly incorporates the synthesized appearance. This approach significantly mitigates structural drift and identity inconsistencies often observed in video editing tasks, leading to spatially coherent synthesis with sharp hair boundaries and well-preserved facial and background details.

\begin{figure*}[ht]
  \centering
  \includegraphics[width=\textwidth]{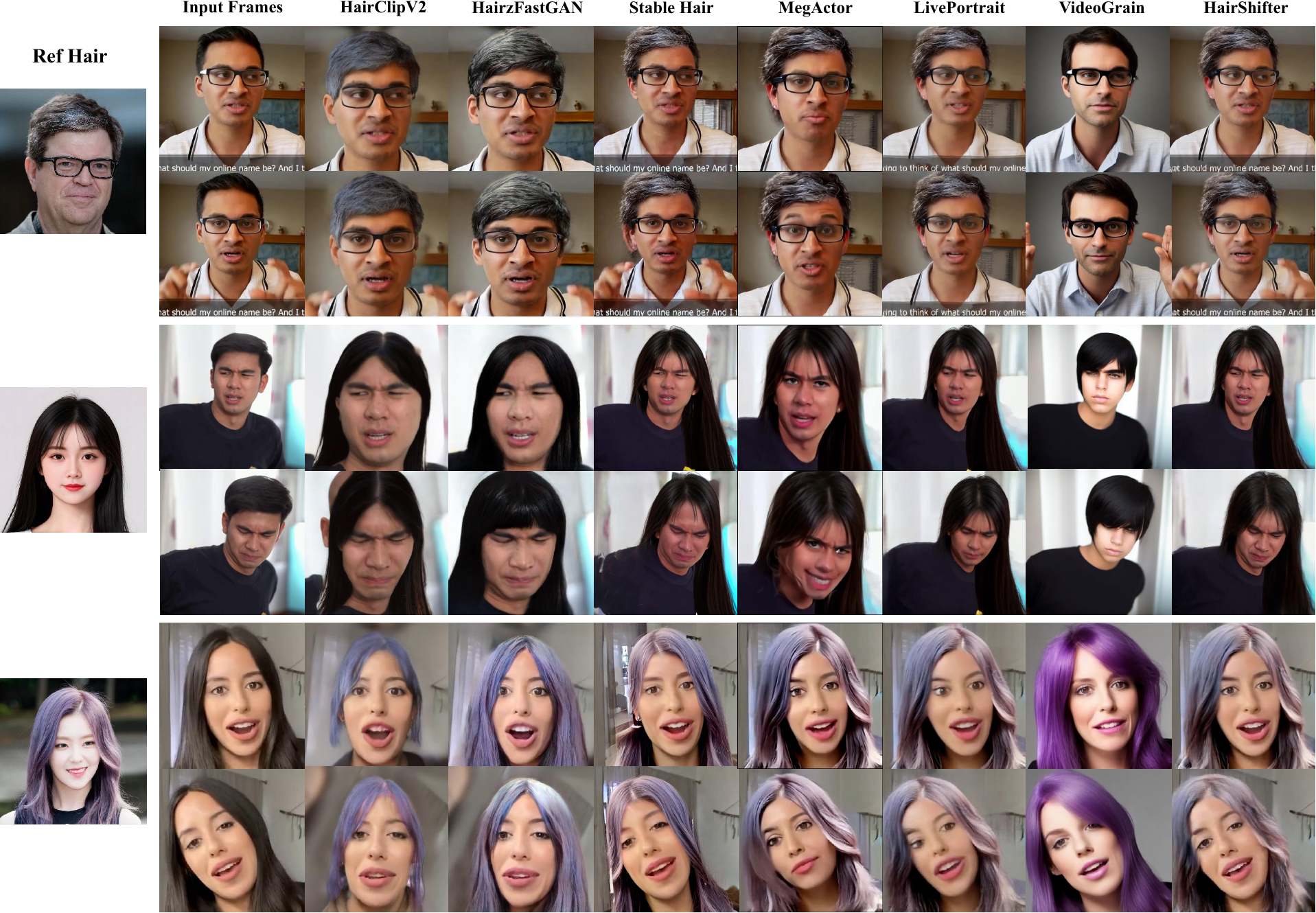}
  \caption{Qualitative Comparison with other methods. HairShifter achieves more refined and stable hairstyle transfer while consistently preserving the details of irrelevant regions between input frames}
  \label{fig:quan}
\end{figure*}

\begin{table*}[t]
  \centering
  \resizebox{\textwidth}{!}{%
    \begin{tabular}{l|cc|cccc|cccc}
      \toprule
      \multirow{2}{*}{Method} 
      & \multicolumn{2}{c|}{\textbf{Video Quality}} 
      & \multicolumn{4}{c|}{\textbf{Non-Hair Region Fidelity}} 
      & \multicolumn{4}{c}{\textbf{Temporal Consistency}} \\
      \cmidrule(lr){2-3} \cmidrule(lr){4-7} \cmidrule(lr){8-11}
      & FID-VID $\downarrow$ & FVD $\downarrow$ 
      & PSNR $\uparrow$ & SSIM $\uparrow$ & L1 $\downarrow$ & IDS $\uparrow$ 
      & Vbench-TF (\%) $\uparrow$ & Vbench-BC (\%) $\uparrow$ & Vbench-MS (\%) $\uparrow$ & CLIP-score $\uparrow$ \\
      \midrule
      HairClipV2\cite{wei2023hairclipv2}   & 62.826          & 1097.600         & \textbf{33.745} & \underline{0.953}          & \underline{0.028}          & \underline{0.735}          & 96.431          & 94.729          & 96.979          & 97.582         \\
      HairFastGAN\cite{nikolaev2024hairfastgan}   & 29.819           & 409.640         & 32.166         & 0.952          & 0.030          & 0.718          & 97.732          & 95.714          & 98.196          & 98.889          \\
      Stable-Hair\cite{zhang2024stable}   & 90.099           & 1161.048         & 31.781         & 0.909 & 0.031 & 0.690 & 94.658          & 92.321          & 95.565          & 95.338          \\
      \midrule
      Megactor\cite{yang2024megactor}      & 36.993           & 367.061          & 30.918         & 0.766          & 0.071          & 0.504          & 98.450          & 95.178          & 99.129          & 99.117          \\
      LivePortrait\cite{guo2024liveportrait}  & \underline{21.007} & \underline{242.575} & 31.062         & 0.821          & 0.055          & 0.568          & \textbf{99.191} & 96.235          & \underline{99.406} & \underline{99.637} \\
      \midrule
      VideoGrain\cite{yang2025videograin}   & 48.286           & 566.635          & 30.076         & 0.824          & 0.071          & 0.363          & 98.506          & \underline{96.481} & 98.989          & 98.946          \\
      \midrule
      \textbf{HairShifter}          & \textbf{11.686}  & \textbf{140.261} & \underline{32.577}   & \textbf{0.959} & \textbf{0.025} & \textbf{0.778} & \underline{99.032} & \textbf{96.501}    & \textbf{99.457}    & \textbf{99.717}    \\
      \bottomrule
    \end{tabular}%
  }
\caption{Quantitative Comparison of Baselines. Bold and underlined scores indicate the 1st- and 2nd-ranked methods.}
\label{tab:quantitative_results}
\end{table*}

\vspace{-0.5em}
\subsection{Training Objectives and Inference Pipeline}
\subsubsection{Optimization Objectives}
We optimize the animation network $G$ using a composite loss function. Following standard practices in portrait animation and generative modeling, similar to frameworks like LivePortrait \cite{guo2024liveportrait}, our objective includes conventional loss terms. These typically comprise an adversarial loss ($\mathcal{L}_{adv}$) to encourage photorealistic outputs, a perceptual loss ($\mathcal{L}_p$) utilizing features from a pre-trained VGG network \cite{simonyan2014very} to enforce high-level feature similarity, and a global L1 loss ($\mathcal{L}_{rec}$) promoting overall structural fidelity between the generated frame $I_p$ and the ground truth frame $I_d$.

To further enhance the preservation and detail within specific regions of interest, we incorporate localized L1 losses that specifically target the hair and non-hair face regions. These losses leverage binary masks, $m_{hair}$ for the hair region and $m_{face}$ for the non-hair face region (excluding hair), which are derived from ground truth facial parsing annotations \cite{zheng2022general}. By focusing optimization on these critical areas, we aim to achieve more precise results. The localized losses $\mathcal{L}_{hair}$ and $\mathcal{L}_{face}$ are defined as:
\begin{equation}
\mathcal{L}_{hair} = \left\| m_{hair} \otimes (I_d - I_p) \right\|_1, \quad \mathcal{L}_{face} = \left\| m_{face} \otimes (I_d - I_p) \right\|_1.
\label{eq:loss_face}
\end{equation}
Here, $\| \cdot \|_1$ signifies the L1 norm, ensuring that errors within these masked areas are directly minimized.

The final objective function guiding the training is a weighted summation of both the standard and these localized refinement loss components:
\begin{equation}
\mathcal{L}_{\text{total}} = \lambda_{adv} \mathcal{L}_{adv} + \lambda_p \mathcal{L}_p + \lambda_{rec} \mathcal{L}_{rec} + \lambda_{hair} \mathcal{L}_{hair} + \lambda_{face} \mathcal{L}_{face}
\label{eq:loss_total}
\end{equation}
where the $\lambda$ coefficients ($\lambda_{adv}, \lambda_p, \lambda_{rec}, \lambda_{hair}, \lambda_{face}$) serve as hyperparameters to balance the relative influence of each loss term during optimization.

\subsubsection{Inference Pipeline}
During inference, given a driving video sequence composed of frames $I_d^t$ where $t\in\{1, ..., T\}$, and a target hairstyle reference image $R_{target}$, the goal is to generate a corresponding output video sequence frame by frame, denoted as $I_p^t$. This process involves two main stages, leveraging the pre-trained IHT module and the disentanglement-trained network $G$. First, a static source image $I_s$ is prepared. An anchor frame $I_{anchor}$ is selected\footnote{{In our practice, we automatically select the source frame with the facial pose most similar to the reference hair image.}} from $V_d$, and the IHT module synthesizes $I_s=\text{IHT}(I_{anchor}, R_{target})$. This image $I_s$ provides the target hair appearance while retaining the identity from $I_{anchor}$. Second, the network $G$ performs frame-by-frame animation. For each driving frame $I_d^t$, $G$ synthesizes the output frame $I_p^t$. It uses $I_s$ as the source for hair appearance and $I_d^t$ to provide the motion cues and the non-hair context (identity, pose, background). Inside, $G$ generates motion-aligned hair features derived from $I_s$ and context features capturing the non-hair regions of $I_d^t$. The core synthesis relies on the Multi-Scale Gated SPADE Decoder $D$, which effectively fuses these two sets of features using its learned gating mechanism (Eq. \ref{eq:gated_fusion}).

This inference strategy capitalizes on the disentanglement learned during training. By processing the static source $I_s$ and the dynamic driving frames $I_d^t$ through complementary pathways within $G$ and fusing them intelligently via the MSG-SPADE Decoder, the framework generates a temporally coherent video with the desired hairstyle, while robustly preserving the subject's identity and non-hair region details.

\vspace{-0.75em}
\section{Experiments}
\subsection{Implementation Detail}
\label{subsec:implementation_detail}

We employ the CelebV-HQ\cite{zhu2022celebv} for model training. Following our decoupled strategy, we use anchor frames $I_s$ and IHT-generated pseudo driving frames ${I_d}^\prime$ to reconstruct the target frame $I_d$. We adapt the LivePortrait \cite{guo2024liveportrait} framework by loading and freezing its pre-trained weights for the motion components ($E_M, E_H, W$), the Context Decoder $D_C$, and the standard SPADE blocks within the Synthesis Decoder $D_S$. Training exclusively focuses on the newly introduced Non-hair Context Encoder $E_C$ and the GF-SPADE blocks integrated into $D_S$. This targeted training runs for 100 epochs on three RTX 3090 GPUs with a batch size of 4 and a learning rate of \(2.0 \times 10^{-5}\). In all of our experiments, the loss weights were set as follows: $\lambda_{adv}=1, \lambda_p=1, \lambda_{rec}=1, \lambda_{hair}=1,$ and $\lambda_{face}=1$. In practice, these weights achieve a well-balanced trade-off across the multiple objectives.

\vspace{-0.5em}
\subsection{Evaluation Protocol}
\subsubsection{Baselines.} We compare our method against several state-of-the-art video generation and editing frameworks, grouped by their underlying paradigm. First, we evaluate against single-image hairstyle transfer models applied frame-by-frame, including \textbf{HairClipV2}\cite{wei2023hairclipv2}, \textbf{HairFastGAN}\cite{nikolaev2024hairfastgan}, and \textbf{Stable-Hair}\cite{zhang2024stable}. Second, we include portrait animation approaches such as \textbf{LivePortrait}\cite{guo2024liveportrait} and \textbf{MegActor}\cite{yang2024megactor}. For fair comparison, these animation models utilize the same high-quality anchor frame generated by our IHT module as input to drive the video synthesis. Finally, we benchmark against recent diffusion-based video editing models \textbf{VideoGrain}\cite{yang2025videograin}. For the text-driven VideoGrain \cite{yang2025videograin}, prompts were generated from reference images using \textit{Google Gemini 2.0 Flash}\cite{team2023gemini}, and the editing region was set to the entire input image. Note that text inherently provides less precise visual control than image references, which may affect fine-grained style replication.

\subsubsection{Metrics}
We evaluate the generated videos using a diverse set of metrics assessing three core aspects: video quality and fidelity, non-hair region preservation, and temporal consistency.
For \textbf{overall video quality and fidelity}, we compute \textbf{FID-VID}\cite{heusel2017gans} to measure the Fréchet distance between Inception feature distributions of generated and source frames, indicating visual quality. We also use \textbf{FVD}\cite{unterthiner2018towards}, which leverages I3D features\cite{carreira2017quo} to assess both appearance and temporal dynamics.
To evaluate \textbf{non-hair region fidelity and identity preservation}, we calculate several metrics specifically on the non-hair areas identified using a face parsing network\cite{zheng2022general}. These include \textbf{PSNR}\cite{sheikh2006statistical} and \textbf{SSIM}\cite{wang2004image} for structural and pixel-level similarity, and \textbf{L1 Loss} for average pixel difference. Additionally, we employ ArcFace\cite{deng2019arcface} to compute \textbf{identity similarity (IDS)} based on facial embeddings, quantifying identity preservation.
For \textbf{temporal consistency}, we adopt relevant metrics from the VBench\cite{huang2024vbench} benchmark: \textbf{Temporal Flicker (TF)}, \textbf{Background Consistency (BC)}, and \textbf{Motion Smoothness (MS)}, where higher percentages indicate better performance for all three. Furthermore, following previous work\cite{wu2023tune}, we compute the average \textbf{CLIP-score\cite{radford2021learning}} measuring cosine similarity between CLIP embeddings of consecutive frames, where higher scores indicate better frame-to-frame visual coherence.

\subsection{Comparison with Baselines}

\textbf{Qualitative Comparison.} Figure~\ref{fig:quan} provides a qualitative comparison using test set videos and social media portraits with distinctive hairstyles as references. Frame-wise methods (HairCLIPV2, HairFastGAN, Stable-Hair) suffer from severe flickering due to poor \textit{spatio-temporal consistency}. HairCLIPV2 and HairFastGAN also exhibit background inconsistencies due to the face alignment preprocessing required by their official implementations. While Stable-Hair achieves good single-frame quality, its inability to consistently adapt the hairstyle under pose variation (e.g., rows 4, 8) highlights the difficulty of achieving \textit{dynamic adaptability} with static models. Portrait animation methods (MegActor, LivePortrait), despite starting from high-quality IHT anchors, struggle with \textit{precise non-hair region preservation}: both exhibit progressive identity drift and background inconsistencies (e.g., hand motion, row 1), revealing the limitation of holistic deformation for targeted replacement tasks, as discussed in the Introduction. VideoGrain performs poorly in both \textit{non-hair region preservation} (altering identity and background details) and achieving fine-grained, reference-driven \textit{high-quality hair transfer}. In contrast, our \textit{HairShifter} 
achieves strong \textit{dynamic adaptability}, robust \textit{spatio-temporal consistency}, and accurate \textit{non-hair region preservation}, even under complex motion and occlusion. These results demonstrate the effectiveness of our "Anchor Frame + Animation" approach, 
particularly the synergy between the IHT anchor generation and the MSG-SPADE Decoder's ability to seamlessly integrate the hair while preserving context.

\noindent\textbf{Quantitative Comparison.} We perform quantitative evaluation on 200 videos from the CelebV-HQ dataset \cite{zhu2022celebv} (100 identities, 100 frames each, outside the training set), using random reference hairs from CelebA-HQ \cite{huang2018introvae}. Table \ref{tab:quantitative_results} presents the results. \textit{HairShifter} achieves state-of-the-art performance across the board, validating its effectiveness in balancing the conflicting demands of video hair transfer.
In terms of \textbf{overall video quality}, our method obtains the best FID-VID and FVD scores, indicating superior visual realism and addressing the challenge of \textit{high-quality generation}.
For \textbf{non-hair region fidelity}, \textit{HairShifter} leads significantly in SSIM, L1 Loss, and IDS, while also achieving a highly competitive PSNR. This quantitatively confirms its success in \textit{precise non-hair region preservation}, directly validating the effectiveness of our decoupled training strategy and the MSG-SPADE Decoder's reference-guided fusion in maintaining identity and background details.
Regarding \textbf{temporal consistency}, our method achieves the highest scores for VBench BC, VBench MS, and CLIP-score, along with a near-best VBench TF. This demonstrates its capability to maintain \textit{spatio-temporal consistency}, mitigating flickering and ensuring smooth transitions, thereby validating the robustness of our "Anchor Frame + Animation" approach. These quantitative results strongly support our qualitative observations and the claims made regarding HairShifter's advancements.

\begin{table}[htbp]
\centering
\resizebox{\columnwidth}{!}{ 
\begin{tabular}{lcccc}
\toprule
\textbf{Method} & \textbf{Accuracy} & \textbf{Preservation} & \textbf{Naturalness} & \textbf{\makecell[c]{Temporal\\Consistency}} \\ 
\midrule
\textbf{HairClipV2}      & 1.340 & 1.769 & 1.184 & 1.211 \\
\textbf{HairFastGAN}     & 1.646 & 2.463 & 1.490 & 1.803 \\
\textbf{Stable-Hair}     & 2.313 & 2.680 & 1.680 & 1.551 \\
\midrule
\textbf{Megactor}        & 3.769 & 3.279 & 3.605 & 3.864 \\
\textbf{LivePortrait}    & 4.136 & 3.442 & 3.864 & 4.224 \\
\midrule
\textbf{VideoGrain}      & 1.490 & 1.224 & 1.714 & 2.109 \\
\midrule
\textbf{HairShifter}            & \textbf{4.361} & \textbf{4.599} & \textbf{4.333} & \textbf{4.463} \\
\bottomrule
\end{tabular}
} 
\caption{User Study on Video Hair Transfer. HairShifter outperforms all others across these four metrics.}
\label{tab:userstudy_transposed}
\end{table}

\noindent\textbf{User Study.} To incorporate human perceptual judgment, we conducted a user study with 20 participants evaluating 30 video triplets (original, reference, generated) across 7 methods. Participants rated results based on four criteria directly related to our core challenges: \textbf{Accuracy} (quality of hair transfer), \textbf{Preservation} (fidelity of non-hair regions), \textbf{Naturalness} (overall realism), and \textbf{Temporal Consistency} (smoothness). As shown in Table \ref{tab:userstudy_transposed}, \textit{HairShifter} was overwhelmingly preferred across all criteria, significantly outperforming competing methods. This confirms that our technical contributions translate into perceptually superior results, successfully balancing generation quality, preservation, and temporal coherence in a way users find most compelling.
\vspace{-0.75em}

\begin{figure}[htbp] 
  \centering
  \includegraphics[width=\columnwidth]{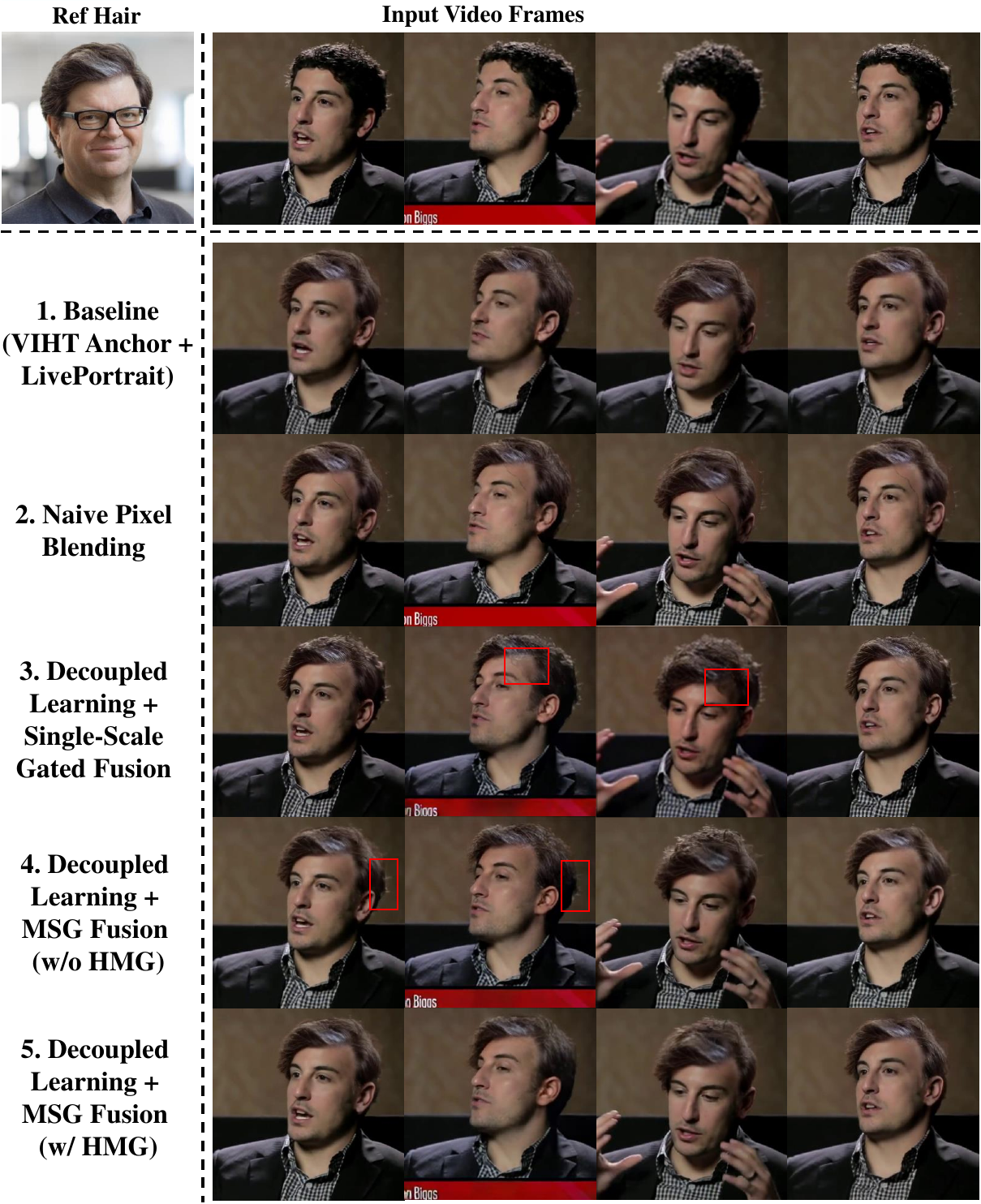}
  \caption{Visual comparison for the Ablation Study. Comparing settings from Baseline (top) to HairShifter (bottom), showcasing progressive gains in blending quality and non-hair region preservation.}
  \label{fig:ablation}
\end{figure}

\begin{table}[htbp]
\centering
\setlength{\tabcolsep}{2pt} 

\resizebox{1\columnwidth}{!}{ 
\begin{tabular}{|c|l|c|c|c|c|c|c|} 
\hline 
\textbf{\#} & \textbf{\makecell[l]{Gated Fusion}} & \textbf{\makecell[c]{Pixel\\Blending}}
& \textbf{\makecell[c]{FID-VID$\downarrow$}}
& \textbf{\makecell[c]{FVD$\downarrow$}}
& \textbf{\makecell[c]{SSIM$\uparrow$}}
& \textbf{\makecell[c]{IDS$\uparrow$}}
& \textbf{\makecell[c]{Vbench-MS$\uparrow$}} \\
\hline 
1 & N/A & $\times$
& 21.007 & 242.575 & 0.821 & 0.568 & 99.406 \\
\hline 
2 & N/A & $\checkmark$ 
&  15.011      &   183.667      &  0.957      &   0.740      &   99.023      \\
\hline 
3 & Single-scale & $\checkmark$
& 14.801 & 146.743 & 0.934 & 0.769 & 99.131 \\
\hline 
4 & Multi-scale w/o HMG & $\checkmark$
& 12.299 & 140.534 & 0.940 & 0.770 & 99.337 \\
\hline 
5 & Multi-scale w/ HMG & $\checkmark$
& \textbf{11.686} & \textbf{140.261} & \textbf{0.959} & \textbf{0.778} & \textbf{99.457} \\
\hline 
\end{tabular}
} 
\caption{Quantitative Ablation Study. Results demonstrate the progressive contribution of each component, with HairShifter(Setting 5) achieving the best overall performance.}
\label{tab:ablation}
\end{table}

\subsection{Ablation Study}

To validate the effectiveness of our key contributions – the decoupled training strategy and the Multi-Scale Gated SPADE (MSG-SPADE) Decoder with Hair Mask Guidance (HMG) – we conduct an ablation study. We evaluate five settings, progressively incorporating our proposed components. The \textbf{Baseline (Setting 1)} employs the standard LivePortrait \cite{guo2024liveportrait} framework driven by our IHT anchor. \textbf{Setting 2 (+ Naive Pixel Blending)} then applies post-processing: it composites the synthesized hair from the Setting 1's predicted frame onto the original driving frame, using a hair mask softened at the edges with Gaussian blur. \textbf{Setting 3 (+ Decoupled Learning \& Single-Scale Fusion)} then introduces our decoupled training strategy and applies learned gated fusion (Eq. \ref{eq:gated_fusion}) at a single decoder scale. Building on this, \textbf{Setting 4 (+ Multi-Scale Fusion w/o HMG)} extends the gated fusion across multiple scales but omits Hair Mask Guidance (HMG) during feature modulation in Eq. \ref{eq:fn} (use $f_n^\prime = f_c$ instead of $f_n$). Finally, \textbf{Setting 5 (+ HMG)} represents our complete model, utilizing multi-scale gated fusion with HMG (using $f_n$ in Eq. \ref{eq:fn}).

The quantitative results in Table \ref{tab:ablation} and qualitative comparisons in Figure \ref{fig:ablation} clearly demonstrate the benefits of each added component. The Baseline (Setting 1) suffers from poor non-hair region fidelity (low SSIM/IDS). While Naive Blending (Setting 2) enforces pixel preservation, it leads to unnatural artifacts and degrades overall quality (worse FID/FVD), highlighting the necessity for learned integration. Introducing single-scale gated fusion with decoupled learning (Setting 3) improves integration and fidelity over naive blending. Extending fusion to multiple scales (Setting 4) significantly enhances spatial coherence and temporal stability (better FID/FVD, VBench-MS vs. Setting 3). Crucially, the comparison between Setting 4 and Setting 5 isolates the impact of HMG. Adding HMG yields substantial improvements in non-hair fidelity (highest SSIM/IDS) and overall realism (lowest FID/FVD), this component fulfills its specific purpose of providing explicit spatial cues for boundary refinement, leading to visibly sharper results. This systematic improvement validates our approach, proving the effectiveness of combining decoupled learning with mask-aware multi-scale gated fusion for achieving identity-preserving, spatially coherent, and temporally stable video hair transfer.

\vspace{-0.5em}
\subsection{Limitations and Discussions}
\begin{figure}[h] 
  \centering
  \includegraphics[width=\columnwidth]{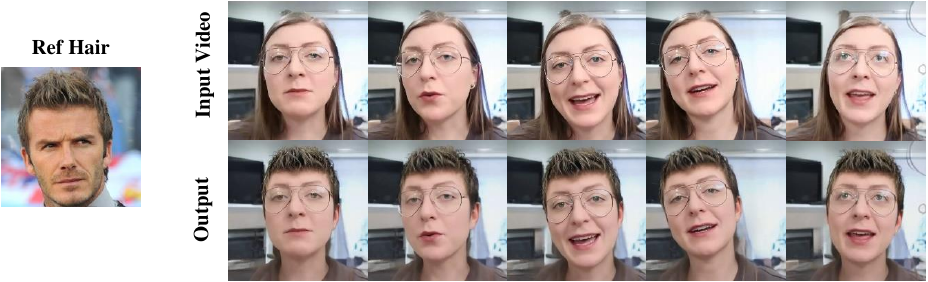}
  \caption{Example limitation showing artifacts when transferring long hair onto a short-haired subject.}
  \label{fig:limit}
\end{figure}

As shown in Fig \ref{fig:limit} , although our method can inherently handle various hairstyles, the hair transfer from long hair into short hair still remains challenging. This relies on the consistent video inpainting for the hair regions. Besides, our current framework primarily bases hair motion on head movement, limiting its ability to simulate secondary dynamics like inertia during rapid rotations. Additionally, complex hand-hair interactions in the driving video can sometimes degrade blending quality near the interaction points. Future work could explore incorporating physics-based modeling or specific interaction handling to address these limitations and enhance realism.
\vspace{-1em}

\vspace{-1.5em}
\section{Conclusion}
We introduced \textit{HairShifter}, the first dedicated framework for video hairstyle transfer. Our "Anchor Frame + Animation" approach integrates high-fidelity anchor generation via Image Hair Transfer(IHT) module with a specialized animation network. Trained using a novel decoupled strategy for effective hair/non-hair disentangle, the network employs our Multi-Scale Gated SPADE (MSG-SPADE) Decoder for seamless blending and precise identity/background preservation. Demonstrating state-of-the-art results, HairShifter successfully balances quality, temporal consistency, and fidelity, establishing a robust baseline for video hairstyle transfer.

\begin{acks}
We thank the anonymous reviewers for their valuable feedback and constructive suggestions. This work was supported by National Natural Science Foundation (Grant No. 62072382) and Yango Charitable Foundation.
\end{acks}

\bibliographystyle{ACM-Reference-Format}
\bibliography{sample-base}

\appendix

\clearpage

\section{Appendix}

\subsection{Adaptation of Image Hair Transfer(IHT)}
The foundation of our video hair transfer pipeline relies on generating a high-fidelity initial frame incorporating the target hairstyle onto the subject's identity. For this purpose, we develop the \textbf{I}mage \textbf{H}air \textbf{T}ransfer (IHT) module, which builds upon the robust Stable-Hair \cite{zhang2024stable} framework, known for its strong diffusion-based image synthesis. A key challenge with the original Stable-Hair lies in its training on single-image derivatives, which inadvertently couples the input face pose with the reference hair pose. This limits its effectiveness when transferring hairstyles between images with differing poses, a common scenario in video. To address this, we improve Stable-Hair by fine-tuning it with video data, specifically using input and reference frames sampled from \textit{different timestamps} within the same sequence. This forces the model to learn pose-invariant hair representations, significantly enhancing its robustness to the pose variations encountered in video applications.

To improve the IHT module, we fine-tune the Hair Extractor and Latent IdentityNet of Stable-Hair, with a combined image-video dataset. For the image portion, we sample 2000 images from FFHQ. Following the methodology of Stable-Hair, we generate training triplets – comprising a \text{GT image}, a \text{Reference image}, and a \text{Bald proxy image} – all derived from one source image. For the video portion, we sample approximately 18,000 frames from CelebV-HQ. For these, the \text{GT image} and \text{Bald proxy image} originate from the same frame while the \text{Reference image} is another frame from the same video, often chosen for pose variation via SixDRepNet\cite{hempel20226d}. This process yields a combined training dataset of over 60,000 triplets. Fine-tuning was conducted for 100,000 steps on three A6000 GPUs, using a batch size of 8 and a learning rate of \(2 \times 10^{-5}\).

\begin{figure}[htbp] 
  \centering
  \includegraphics[width=\linewidth]{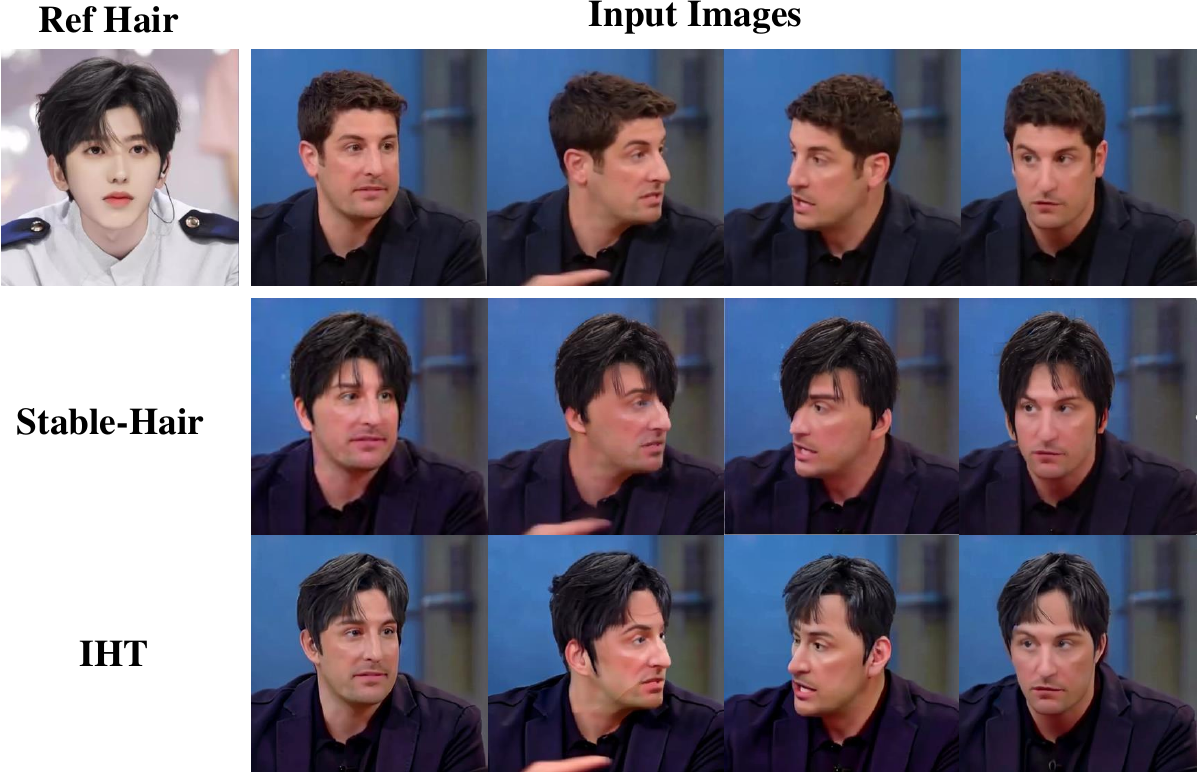} 
  \caption{Comparison of our adapted Image Hair Transfer (IHT) module with the baseline Stable-Hair. Our fine-tuned IHT demonstrates improved robustness, generating a hairstyle more consistent with the input posV Ce compared to the baseline.}
  \label{fig:iht_comparison} 
\end{figure}

As shown in Figure \ref{fig:iht_comparison}, the resulting IHT module demonstrates improved capability in generating spatially coherent and structurally accurate hairstyles, despite substantial pose discrepancies between the anchor frame and the reference hair image. This establishes a stable, high-fidelity foundation for subsequent animation. Within our framework, IHT serves two distinct functions: supporting training by generating pseudo driving frames, and enabling inference by synthesizing the high-quality anchor frame that defines the target hair appearance.

\vspace{-0.5em}
\subsection{Text Prompt Generation for VideoGrain Baseline}
\label{sec:supp_prompt_gen}

To evaluate the text-driven VideoGrain baseline \cite{yang2025videograin} comparably against reference-image-driven methods, we generated descriptive text prompts from the target hair reference images ($R_{target}$). We utilized the Multimodal Large Language Model (MLLM) \textit{Google Gemini 2.0 Flash} \cite{team2023gemini} for this task. Each reference image was input to the MLLM along with the following instruction prompt:
\begin{quote}
\small 
\texttt{Analyze the given facial image and generate a descriptive text embedding. The text should include the following details:} \\ 
\texttt{1. Gender - Clearly specify whether the person appears to be male, female, or another gender identity.} \\
\texttt{2. Hair Color - Describe the hair color, such as black, blonde, brown, red, etc.} \\
\texttt{3. Hair Shape - Specify the shape or style of the hair, such as straight, curly, wavy, short, long, etc.} \\
\texttt{4. Hair Structure - Describe the hair texture and structure, including volume, layering, smoothness, or other defining characteristics.} \\
\texttt{Avoid mentioning the background, environment, or any elements unrelated to the person's facial features. Focus solely on the person's facial and hair attributes.}
\end{quote}

The text description generated by the MLLM for a given reference image, focusing on the hair attributes as requested, was then directly used as the input prompt for VideoGrain when processing the corresponding driving video ($V_d$). For these experiments, the editing region was set to the entire image frame, omitting a specific mask. This approach ensured the textual guidance for VideoGrain was derived from the same visual targets used across all compared methods.

\vspace{-0.5em}
\subsection{Computational Cost and Real-Time Capability}

HairShifter contains approximately 3.45B parameters and sustains \textbf{20+ FPS} on $512 \times 512$ resolution videos using a single RTX 3090 GPU (23 GB VRAM) without any acceleration, significantly surpassing existing methods or alternative solutions that typically operate near \textbf{1 FPS}. Inference requires a one-time invocation of the high-fidelity \textit{IHT module} for the anchor frame, incurring \textbf{74.23 TFLOPs}. All remaining frames are handled by the animation module with only \textbf{1.57 TFLOPs per frame}. The average computational cost per frame over a video of length $N$ is:
\[
C(N) = \frac{74.23 + 1.57 \times N}{N} \text{ TFLOPs},
\]
which quickly approaches $1.57$ as $N$ grows. This amortized formulation ensures that per-frame overhead becomes negligible in longer sequences, enabling \textbf{real-time, high-resolution video editing} with consistent performance.

\end{sloppypar}
\end{document}